\documentclass[rmp,aps,twocolumn,nofootinbib]{revtex4}

\usepackage{pslatex} 
\usepackage{amsmath}
\usepackage[]{graphicx} 

\long\def\symbolfootnote[#1]#2{\begingroup \def\thefootnote{\fnsymbol{footnote}}\footnote[#1]{#2}\endgroup}

\newcommand{\comment}[1]{} 

\begin{document}

\title{A Tutorial on Principal Component Analysis}
\date{\today; Version 3.02}
\author{Jonathon Shlens} 
\email{jonathon.shlens@gmail.com}
\affiliation{
Google Research \\
Mountain View, CA 94043}

\begin{abstract}
Principal component analysis (PCA) is a mainstay of modern data analysis - a black box that is widely used but (sometimes) poorly understood. The goal of this paper is to dispel the magic behind this black box. This manuscript focuses on building a solid intuition for how and why principal component analysis works. This manuscript crystallizes this knowledge by deriving from simple intuitions, the mathematics behind PCA. This tutorial does not shy away from explaining the ideas informally, nor does it shy away from the mathematics. The hope is that by addressing both aspects, readers of all levels will be able to gain a better understanding of PCA as well as the when, the how and the why of applying this technique.
\end{abstract}

\maketitle

\section{Introduction}
Principal component analysis (PCA) is a standard tool in modern data analysis - in diverse fields from neuroscience to computer graphics - because it is a simple, non-parametric method for extracting relevant information from confusing data sets. With minimal effort PCA provides a roadmap for how to reduce a complex data set to a lower dimension to reveal the sometimes hidden, simplified structures that often underlie it.
 
The goal of this tutorial is to provide both an intuitive feel for PCA, and a thorough discussion of this topic. We will begin with a simple example and provide an intuitive explanation of the goal of PCA. We will continue by adding mathematical rigor to place it within the framework of linear algebra to provide an explicit solution.  We will see how and why PCA is intimately related to the mathematical technique of singular value decomposition (SVD). This understanding will lead us to a prescription for how to apply PCA in the real world and an appreciation for the underlying assumptions. My hope is that a thorough understanding of PCA provides a foundation for approaching the fields of machine learning and dimensional reduction.

The discussion and explanations in this paper are informal in the spirit of a tutorial. The goal of this paper is to {\it educate}. Occasionally, rigorous mathematical proofs are necessary although relegated to the Appendix. Although not as vital to the tutorial, the proofs are presented for the adventurous reader who desires a more complete understanding of the math. My only assumption is that the reader has a working knowledge of linear algebra. My goal is to provide a thorough discussion by largely building on ideas from linear algebra and avoiding challenging topics in statistics and optimization theory (but see Discussion). Please feel free to contact me with any suggestions, corrections or comments.

\section{Motivation: A Toy Example}

Here is the perspective: we are an experimenter. We are trying to understand some phenomenon by measuring various quantities (e.g. spectra, voltages, velocities, etc.) in our system. Unfortunately, we can not figure out what is happening because the data appears clouded, unclear and even redundant. This is not a trivial problem, but rather a fundamental obstacle in empirical science. Examples abound from complex systems such as neuroscience, web indexing, meteorology and oceanography - the number of variables to measure can be unwieldy and at times even {\it deceptive}, because the underlying relationships can often be quite simple.

\begin{figure}[t]
\centering
\includegraphics[width=0.47\textwidth]{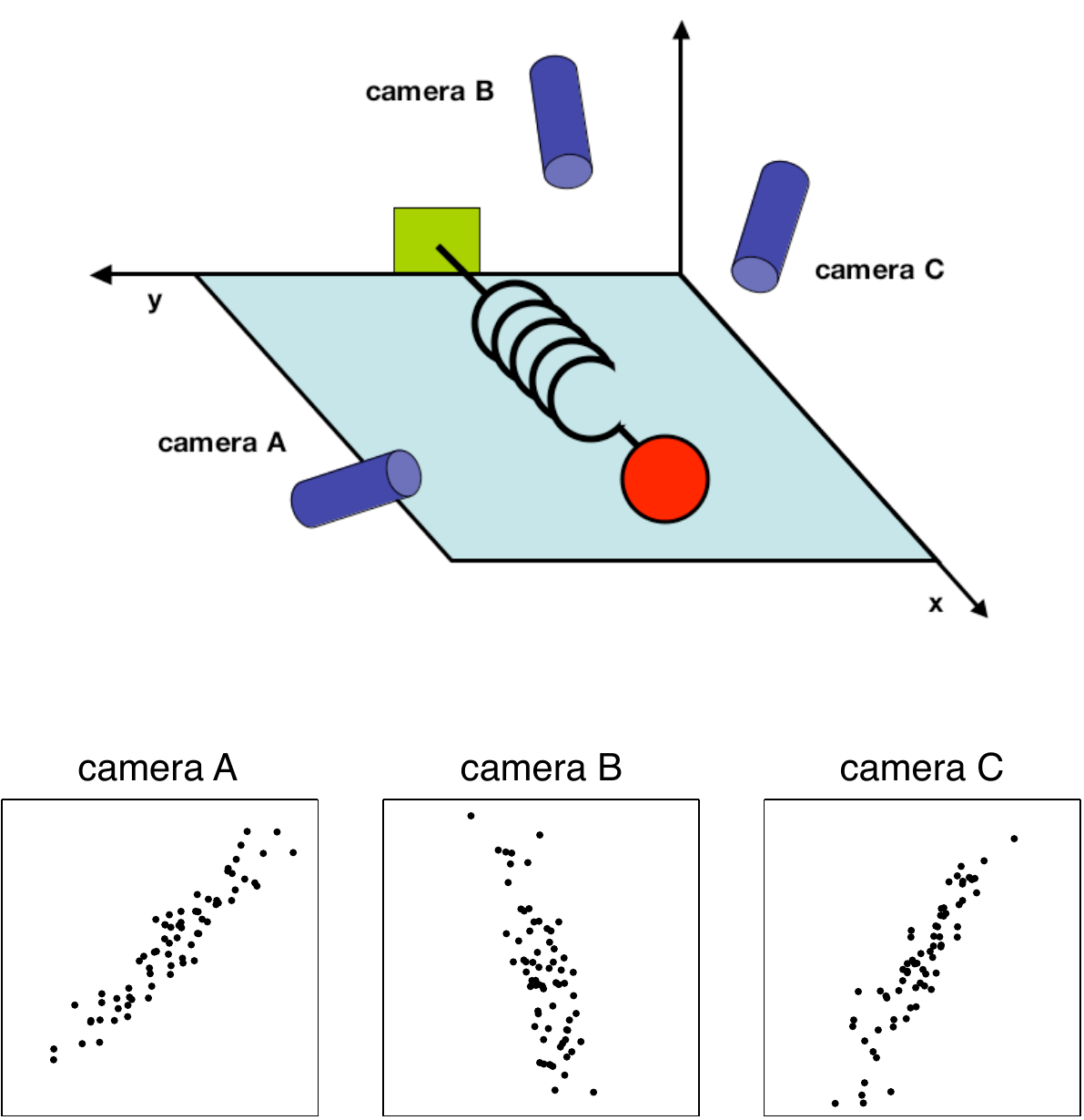}
\caption{A toy example. The position of a ball attached to an oscillating spring is recorded using three cameras A, B and C. The position of the ball tracked by each camera is depicted in each panel below.}
\label{diagram:toy}
\end{figure}

Take for example a simple toy problem from physics diagrammed in Figure~\ref{diagram:toy}. Pretend we are studying the motion of the physicist's ideal spring. This system consists of a ball of mass $m$ attached to a massless, frictionless spring. The ball is released a small distance away from equilibrium (i.e. the spring is stretched). Because the spring is ideal, it oscillates indefinitely along the $x$-axis about its equilibrium at a set frequency.

This is a standard problem in physics in which the motion along the $x$ direction is solved by an explicit function of time. In other words, the underlying dynamics can be expressed as a function of a single variable $x$.

However, being ignorant experimenters we do not know any of this. We do not know which, let alone how many, axes and dimensions are important to measure. Thus, we decide to measure the ball's position in a three-dimensional space (since we live in a three dimensional world). Specifically, we place three movie cameras around our system of interest. At \mbox{120 Hz} each movie camera records an image indicating a two dimensional position of the ball (a projection). Unfortunately, because of our ignorance, we do not even know what are the real $x$, $y$ and $z$ axes, so we choose three camera positions $\vec{\mathbf{a}}, \vec{\mathbf{b}}$ and $\vec{\mathbf{c}}$ at some arbitrary angles with respect to the system. The angles between our measurements might not even be $90^{o}$! Now, we record with the cameras for several minutes. The big question remains: {\it how do we get from this data set to a simple equation of $x$?}

We know a-priori that if we were smart experimenters, we would have just measured the position along the $x$-axis with one camera. But this is not what happens in the real world. We often do not know which measurements best reflect the dynamics of our system in question. Furthermore, we sometimes record more dimensions than we actually need.

Also, we have to deal with that pesky, real-world problem of noise. In the toy example this means that we need to deal with air, imperfect cameras or even friction in a less-than-ideal spring. Noise contaminates our data set only serving to obfuscate the dynamics further. {\it This toy example is the challenge experimenters face everyday.} Keep this example in mind as we delve further into abstract concepts. Hopefully, by the end of this paper we will have a good understanding of how to systematically extract $x$ using principal component analysis.

\section {Framework: Change of Basis}

The goal of principal component analysis is to identify the most meaningful basis to re-express a data set. The hope is that this new basis will filter out the noise and reveal hidden structure. In the example of the spring, the explicit goal of PCA is to determine: ``the dynamics are along the $x$-axis.'' In other words, the goal of PCA is to determine that $\hat{\mathbf{x}}$, i.e. the unit basis vector along the $x$-axis, is the important dimension. Determining this fact allows an experimenter to discern which dynamics are important, redundant or noise.

\subsection{A Naive Basis}
With a more precise definition of our goal, we need a more precise definition of our data as well. We treat every time sample (or experimental trial) as an individual sample in our data set. At each time sample we record a set of data consisting of multiple measurements (e.g. voltage, position, etc.). In our data set, at one
point in time, camera {\it A} records a corresponding ball position $\left(x_A,y_A\right)$. One sample or trial can then be expressed as a 6 dimensional column vector
$$\vec{X}= \left[ \begin{array}{c} x_A \\ y_A \\ x_B \\ y_B \\ x_C \\ y_C \\ \end{array} \right]$$
where each camera contributes a 2-dimensional projection of the ball's position to the entire vector $\vec{X}$. If we record the ball's position for 10 minutes at 120 Hz, then we have recorded $10\times 60 \times 120 = 72000 $ of these vectors.

With this concrete example, let us recast this problem in abstract terms.  Each sample $\vec{X}$ is an $m$-dimensional vector, where $m$ is the number of measurement types. Equivalently, every sample is a vector that lies in an $m$-dimensional vector space spanned by some orthonormal basis. From linear algebra we know that all measurement vectors form a linear combination of this set of unit length basis vectors. What is this orthonormal basis?

This question is usually a tacit assumption often overlooked. Pretend we gathered our toy example data above, but only looked at camera $A$. What is an orthonormal basis for $(x_A, y_A)$?  A naive choice would be $\{(1,0),(0,1)\}$, but why select this basis over $\{ (\frac{\sqrt{2}}{2},\frac{\sqrt{2}}{2}), (\frac{-\sqrt{2}}{2},\frac{-\sqrt{2}}{2})\}$ or any other arbitrary rotation?  The reason is that the {\it naive basis reflects the method we gathered the data.} Pretend we record the position $(2,2)$. We did not record $2\sqrt{2}$ in the $(\frac{\sqrt{2}}{2},\frac{\sqrt{2}}{2})$ direction and $0$ in the perpendicular direction. Rather, we recorded the position $(2,2)$ on our camera meaning 2 units up and 2 units to the left in our camera window. Thus our original basis reflects the method we measured our data.

How do we express this naive basis in linear algebra? In the two dimensional case, $\{(1,0),(0,1)\}$ can be recast as individual row vectors. A matrix constructed out of these row vectors is the $2\times 2$ identity matrix $I$. We can generalize this to the $m$-dimensional case by constructing an $m\times m$ identity matrix
$$\mathbf{B} = \left[ \begin{array}{c} \mathbf{b_1} \\ \mathbf{b_2} \\ \vdots \\ \mathbf{b_m} \end{array} \right] = 
\left[ \begin{array}{cccc} 1 & 0 & \cdots & 0 \\ 0 & 1 & \cdots & 0 \\ \vdots & \vdots & \ddots & \vdots \\ 0 & 0 & \cdots & 1 \\ \end{array} \right] = \mathbf{I}$$
where each {\it row} is an orthornormal basis vector $\mathbf{b}_i$ with $m$ components. We can consider our naive basis as the effective starting point. All of our data has been recorded in this basis and thus it can be trivially expressed as a linear combination of $\{\mathbf{b}_i\}$.

\subsection{Change of Basis}
With this rigor we may now state more precisely what PCA asks: {\it Is there another basis, which is a linear combination of the original basis, that best re-expresses our data set?}

A close reader might have noticed the conspicuous addition of the word {\it linear}. Indeed, PCA makes one stringent but powerful assumption: linearity. Linearity vastly simplifies the problem by restricting the set of potential bases. With this assumption PCA is now limited to re-expressing the data as a {\it linear combination} of its basis vectors. 

Let $\mathbf{X}$ be the original data set, where each $column$ is a single sample (or moment in time) of our data set (i.e. $\vec{X}$). In the toy example $\mathbf{X}$ is an $m\times n$ matrix where $m=6$ and $n=72000$.  Let $\mathbf{Y}$ be another $m \times n$ matrix related by a linear transformation $\mathbf{P}$. $\mathbf{X}$ is the original recorded data set and $\mathbf{Y}$ is a new representation of that data set.
\begin{equation}
\mathbf{PX = Y}
\label{eqn:basis-transform}
\end{equation}
Also let us define the following quantities.\footnote{In this section $\mathbf{x_i}$ and $\mathbf{y_i}$ are {\it column} vectors, but be forewarned. In all other sections $\mathbf{x_i}$ and $\mathbf{y_i}$ are {\it row} vectors.}
\begin{itemize}
\item $\mathbf{p_i}$ are the rows of $\mathbf{P}$
\item $\mathbf{x_i}$ are the columns of $\mathbf{X}$ (or individual $\vec{X})$.
\item $\mathbf{y_i}$ are the columns of $\mathbf{Y}$.
\end{itemize}
Equation~\ref{eqn:basis-transform} represents a change of basis and thus can have many interpretations.
\begin{enumerate}
\item $\mathbf{P}$ is a matrix that transforms $\mathbf{X}$ into $\mathbf{Y}$.
\item Geometrically, $\mathbf{P}$ is a rotation and a stretch which again transforms $\mathbf{X}$ into $\mathbf{Y}$.
\item The rows of $\mathbf{P}$, $\left\{\mathbf{p_1}, \ldots, \mathbf{p_m}\right\}$, are a set of new basis vectors for expressing the columns of $\mathbf{X}$.
\end{enumerate}
The latter interpretation is not obvious but can be seen by writing out the explicit dot products of $\mathbf{PX}$.
\begin{eqnarray*}
\mathbf{PX} & = & \left[ \begin{array}{c} \mathbf{p_1} \\ \vdots \\ \mathbf{p_m} \\ \end{array} \right] \left[ \begin{array}{ccc} \mathbf{x_1} & \cdots & \mathbf{x_n} \\ \end{array} \right] \\
\mathbf{Y}  & = & \left[ \begin{array}{ccc} \mathbf{p_1 \cdot x_1} & \cdots & \mathbf{p_1 \cdot x_n} \\ \vdots & \ddots & \vdots \\ \mathbf{p_m \cdot x_1} & \cdots & \mathbf{p_m \cdot x_n} \\ \end{array} \right] \\  
\end{eqnarray*}
We can note the form of each column of $\mathbf{Y}$.
$$\mathbf{y}_i = \left[ \begin{array}{c} \mathbf{p_1\cdot x_i} \\ \vdots \\ \mathbf{p_m\cdot x_i} \\ \end{array}\right]$$
We recognize that each coefficient of $\mathbf{y_i}$ is a dot-product of $\mathbf{x_i}$ with the corresponding row in $\mathbf{P}$. In other words, the $j^{th}$ coefficient of $\mathbf{y_i}$ is a projection on to the $j^{th}$ row of $\mathbf{P}$. This is in fact the very form of an equation where $\mathbf{y_i}$ is a projection on to the basis of $\left\{\mathbf{p_1}, \ldots, \mathbf{p_m}\right\}$. Therefore, the rows of $\mathbf{P}$ are a new set of basis vectors for
representing of columns of $\mathbf{X}$.

\subsection{Questions Remaining}
By assuming linearity the problem reduces to finding the appropriate {\it change of basis}. The row vectors $\left\{\mathbf{p_1}, \ldots, \mathbf{p_m}\right\}$ in this transformation will become the {\it principal components} of $\mathbf{X}$. Several questions now arise.
\begin{itemize}
\item What is the best way to re-express $\mathbf{X}$? 
\item What is a good choice of basis $\mathbf{P}$? 
\end{itemize}
These questions must be answered by next asking ourselves {\it what features we would like $\mathbf{Y}$ to exhibit}. Evidently, additional assumptions beyond linearity are required to arrive at a reasonable result. The selection of these assumptions is the subject of the next section.

\section{Variance and the Goal}
Now comes the most important question: what does {\it best express} the data mean? This section will build up an intuitive answer to this question and along the way tack on additional assumptions. 

\subsection{Noise and Rotation}

\begin{figure}
\includegraphics[width=0.25\textwidth]{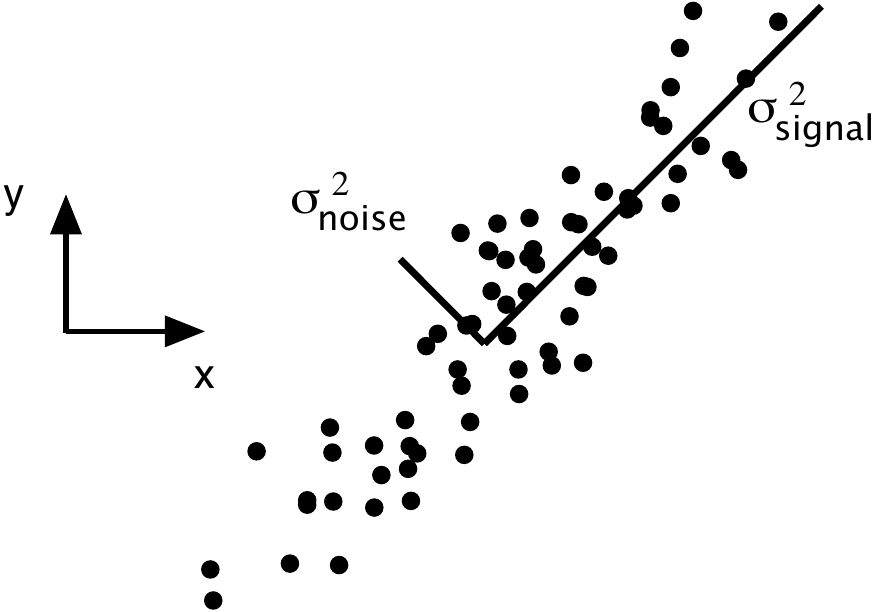}
\caption{Simulated data of $(x, y)$ for camera {\it A}. The signal and noise variances $\sigma_{signal}^{2}$ and $\sigma_{noise}^{2}$ are graphically represented by the two lines subtending the cloud of data. Note that the largest direction of variance does not lie along the basis of the recording $(x_A, y_A)$ but rather along the best-fit line.}
\label{fig:snr}
\end{figure}

Measurement noise in any data set must be low or else, no matter the analysis technique, no information about a signal can be extracted. There exists no absolute scale for noise but rather all noise is quantified relative to the signal strength. A common measure is the {\it signal-to-noise ratio} ({\it SNR}), or a ratio of variances $\sigma^2$,
$$SNR = \frac{\sigma_{signal}^{2}}{\sigma_{noise}^{2}}.$$
A high SNR ($\gg 1$) indicates a high precision measurement, while a low SNR indicates very noisy data.

Let's take a closer examination of the data from camera {\it A} in Figure~\ref{fig:snr}. Remembering that the spring travels in a straight line, every individual camera should record motion in a straight line as well. Therefore, any spread deviating from straight-line motion is noise. The variance due to the signal and noise are indicated by each line in the diagram. The ratio of the two lengths measures how skinny the cloud is: possibilities include a thin line (SNR $\gg 1$), a circle (SNR $ = 1$) or even worse.  By positing reasonably good measurements,  quantitatively we assume that directions with largest variances in our measurement space contain the dynamics of interest. In Figure~\ref{fig:snr} the direction with the largest variance is not $\hat{x}_A = (1,0)$ nor $\hat{y}_A = (0,1)$, but the direction along the long axis of the cloud. Thus, by assumption the dynamics of interest exist along directions with largest variance and presumably highest SNR.

Our assumption suggests that the basis for which we are searching is not the naive basis because these directions (i.e. $(x_A, y_A)$) do not correspond to the directions of largest variance. Maximizing the variance (and by assumption the SNR)  corresponds to finding the appropriate rotation of the naive basis. This intuition corresponds to finding the direction indicated by the line $\sigma^2_{signal}$ in Figure~\ref{fig:snr}. In the 2-dimensional case of Figure~\ref{fig:snr} the direction of largest variance corresponds to the best-fit line for the data cloud.  Thus, rotating the naive basis to lie parallel to the best-fit line would reveal the direction of motion of the spring for the 2-D case. How do we generalize this notion to an arbitrary number of dimensions? Before we approach this question we need to examine this issue from a second perspective.

\subsection{Redundancy}

\begin{figure}
\includegraphics[width=0.47\textwidth]{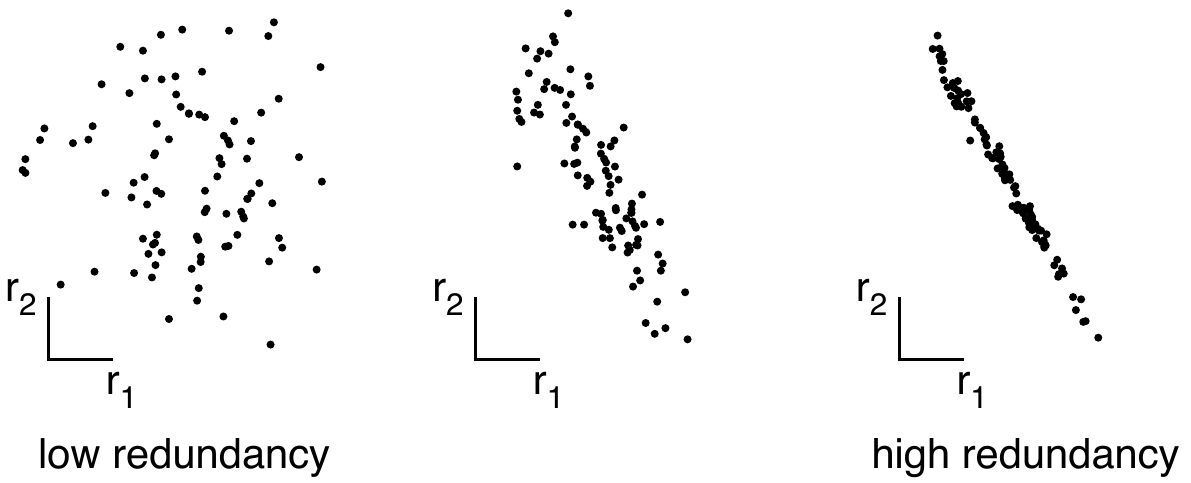}
\caption{A spectrum of possible redundancies in data from the two separate measurements $r_1$ and $r_2$. The two measurements on the left are uncorrelated because one can not predict one from the other. Conversely, the two measurements on the right are highly correlated indicating highly redundant measurements.}
\label{fig:redundancy}
\end{figure}

Figure~\ref{fig:snr} hints at an additional confounding factor in our data - redundancy. This issue is particularly evident in the example of the spring. In this case multiple sensors record the same dynamic information. Reexamine Figure~\ref{fig:snr} and ask whether it was really necessary to record 2 variables. Figure~\ref{fig:redundancy} might reflect a range of possibile plots between two arbitrary measurement types $r_1$ and $r_2$. The left-hand panel depicts two recordings with no apparent relationship.  Because one can not predict $r_1$ from $r_2$, one says that $r_1$ and $r_2$ are uncorrelated.

On the other extreme, the right-hand panel of Figure~\ref{fig:redundancy} depicts highly correlated recordings. This extremity might be achieved by several means:
\begin{itemize}
\item A plot of $(x_A,x_B)$ if cameras {\it A} and {\it B} are very nearby.
\item A plot of $(x_A,\tilde{x}_A)$ where $x_A$ is in meters and $\tilde{x}_A$ is in inches.
\end{itemize}
Clearly in the right panel of Figure~\ref{fig:redundancy} it would be more meaningful to just have recorded a single variable, not both. Why? Because one can calculate $r_1$ from $r_2$ (or vice versa) using the best-fit line. Recording solely one response would express the data more concisely and reduce the number of sensor recordings ($2\rightarrow 1$ variables). Indeed, this is the central idea behind dimensional reduction.

\subsection{Covariance Matrix}
In a 2 variable case it is simple to identify redundant cases by finding the slope of the best-fit line and judging the quality of the fit. How do we quantify and generalize these notions to arbitrarily higher dimensions? Consider two sets of measurements with zero means
$$A = \left\{a_1,a_2,\ldots,a_n\right\}\;,\;\;\;B=\left\{b_1,b_2,\ldots,b_n\right\}$$
where the subscript denotes the sample number. The variance of $A$ and $B$ are individually defined as,
$$\sigma^{2}_{A} = \frac{1}{n}\sum_i a^2_i, \;\;\;\; \sigma^{2}_{B} = \frac{1}{n}\sum_i b^2_i$$
The {\it covariance} between $A$ and $B$ is a straight-forward generalization.
$$covariance\;of\;A\;and\;B \equiv \sigma^{2}_{AB} = \frac{1}{n} \sum_i a_i b_i$$
The covariance measures the degree of the linear relationship between two variables. A large positive value indicates positively correlated data. Likewise, a large negative value denotes negatively correlated data. The absolute magnitude of the covariance measures the degree of redundancy. Some additional facts about the covariance.
\begin{itemize}
\item $\sigma_{AB}$ is zero if and only if $A$ and $B$ are uncorrelated (e.g. Figure~\ref{fig:snr}, left panel).
\item $\sigma_{AB}^{2} = \sigma_{A}^{2}$ if $A=B$.
\end{itemize}
We can equivalently convert $A$ and $B$ into corresponding row vectors.
\begin{eqnarray*}
\mathbf{a} & = & \left[a_1\;a_2\;\ldots\;a_n\right] \\
\mathbf{b} & = & \left[b_1\;b_2\;\ldots\;b_n\right]
\end{eqnarray*}
so that we may express the covariance as a dot product matrix computation.\footnote{Note that in practice, the covariance $\sigma^2_{AB}$ is calculated as $\frac{1}{n-1}\sum_i a_i b_i$. The slight change in normalization constant arises from estimation theory, but that is beyond the scope of this tutorial.}
\begin{equation}
\sigma^{2}_{\mathbf{ab}} \equiv \frac{1}{n} \mathbf{ab}^T
\label{eqn:value-covariance}
\end{equation}

Finally, we can generalize from two vectors to an arbitrary number. Rename the row vectors $\mathbf{a}$ and $ \mathbf{b}$ as $\mathbf{x_1}$ and $\mathbf{x_2}$, respectively, and consider additional indexed row vectors $\mathbf{x_3},\ldots,\mathbf{x_m}$. Define a new $m \times n$ matrix $\mathbf{X}$.
$$\mathbf{X} = \left[ \begin{array}{c} \mathbf{x_1} \\ \vdots \\ \mathbf{x_m} \\ \end{array} \right]$$
One interpretation of $\mathbf{X}$ is the following. Each {\it row} of $\mathbf{X}$ corresponds to all measurements of a particular type. Each {\it column} of $\mathbf{X}$ corresponds to a set of measurements from one particular trial (this is $\vec{X}$ from section 3.1). We now arrive at a definition for the {\it covariance matrix} $\mathbf{C_X}$.
$$\mathbf{C_X} \equiv \frac{1}{n}\mathbf{X}\mathbf{X}^T.$$
Consider the matrix $\mathbf{C_X} = \frac{1}{n}\mathbf{XX}^{T}$. The $ij^{th}$ element of $\mathbf{C_X}$ is the dot product between the vector of the $i^{th}$ measurement type with the vector of the $j^{th}$ measurement type. We can summarize several properties of $\mathbf{C_X}$:
\begin{itemize}
\item $\mathbf{C_X}$ is a square symmetric $m \times m$ matrix (Theorem 2 of Appendix A)
\item The diagonal terms of $\mathbf{C_X}$ are the {\it variance} of particular measurement types.
\item The off-diagonal terms of $\mathbf{C_X}$ are the {\it covariance} between measurement types.
\end{itemize}
$\mathbf{C_X}$ captures the covariance between all possible pairs of measurements. The covariance values reflect the noise and redundancy in our measurements.
\begin{itemize}
\item In the diagonal terms, by assumption, large values correspond to interesting structure.
\item In the off-diagonal terms large magnitudes correspond to high redundancy.
\end{itemize}
Pretend we have the option of manipulating $\mathbf{C_X}$. We will suggestively define our manipulated covariance matrix $\mathbf{C_Y}$. What features do we want to optimize in $\mathbf{C_Y}$?

\subsection{Diagonalize the Covariance Matrix}
We can summarize the last two sections by stating that our goals are (1) to minimize redundancy, measured by the magnitude of the covariance, and (2) maximize the signal, measured by the variance.  What would the optimized covariance matrix $\mathbf{C_Y}$ look like? 
\begin{itemize}
\item All off-diagonal terms in $\mathbf{C_Y}$ should be zero. Thus, $\mathbf{C_Y}$ must be a diagonal matrix. Or, said another way, $\mathbf{Y}$ is decorrelated.
\item Each successive dimension in $\mathbf{Y}$ should be rank-ordered according to variance.
\end{itemize}
There are many methods for diagonalizing $\mathbf{C_Y}$. It is curious to note that PCA arguably selects the easiest method: PCA assumes that all basis vectors $\left\{\mathbf{p_1}, \ldots, \mathbf{p_m}\right\}$ are orthonormal, i.e. $\mathbf{P}$ is an {\it orthonormal matrix}. Why is this assumption easiest?

Envision how PCA works. In our simple example in Figure~\ref{fig:snr}, $\mathbf{P}$ acts as a generalized rotation to align a basis with the axis of maximal variance. In multiple dimensions this could be performed by a simple algorithm: 
\begin{enumerate}
\item Select a normalized direction in $m$-dimensional space along which the variance in $\mathbf{X}$ is maximized. Save this vector as $\mathbf{p_1}$.
\item Find another direction along which variance is maximized, however, because of the orthonormality condition, restrict the search to all directions orthogonal to all previous selected directions. Save this vector as $\mathbf{p_i}$
\item Repeat this procedure until $m$ vectors are selected.
\end{enumerate}
The resulting ordered set of $\mathbf{p}$'s are the {\it principal
components}.

In principle this simple algorithm works, however that would bely the true reason why the orthonormality assumption is judicious. The true benefit to this assumption is that there exists an efficient, analytical solution to this problem. We will discuss two solutions in the following sections.
 
Notice what we gained with the stipulation of rank-ordered variance. We have a method for judging the importance of the principal direction. Namely, the variances associated with each direction $\mathbf{p_i}$ quantify how ``principal'' each direction is by rank-ordering each basis vector $\mathbf{p_i}$ according to the corresponding variances.We will now pause to review the implications of all the assumptions made to arrive at this mathematical goal.

\subsection{Summary of Assumptions}
This section provides a summary of the assumptions behind PCA and hint at when these assumptions might perform poorly.

\newcounter{assumptions-count}
\begin{list}{{\bf \Roman{assumptions-count}}.}
	{\usecounter{assumptions-count}
	\setlength{\rightmargin}{\leftmargin}}
	
\item {\it Linearity}
\\
Linearity frames the problem as a change of basis. Several areas of research have explored how extending these notions to nonlinear regimes (see Discussion).

\item {\it Large variances have important structure.}
\\
This assumption also encompasses the belief that the data has a high SNR. Hence, principal components with larger associated variances represent interesting structure, while those with lower variances represent noise. Note that this is a strong, and sometimes, incorrect assumption (see Discussion).

\item {\it The principal components are orthogonal.}
\\
This assumption provides an intuitive simplification that makes PCA soluble with linear algebra decomposition techniques. These techniques are highlighted in the two following sections.
\end{list}

We have discussed all aspects of deriving PCA - what remain are the linear algebra solutions. The first solution is somewhat straightforward while the second solution involves understanding an important algebraic decomposition.

\section{Solving PCA Using Eigenvector Decomposition }

We derive our first algebraic solution to PCA based on an important property of eigenvector decomposition. Once again, the data set is $\mathbf{X}$,
an $m \times n$ matrix, where $m$ is the number of measurement types and $n$ is the number of samples. The goal is summarized as follows.
\begin{quote}
Find some orthonormal matrix $\mathbf{P}$ in \mbox{$\mathbf{Y=PX}$} such that \mbox{$\mathbf{C_Y} \equiv \frac{1}{n}\mathbf{Y}\mathbf{Y}^T$} is a diagonal matrix. The rows of $\mathbf{P}$ are the {\it principal components} of $\mathbf{X}$.
\end{quote}
We begin by rewriting $\mathbf{C_Y}$ in terms of the unknown variable.
\begin{eqnarray*}
\mathbf{C_Y} & = & \frac{1}{n}\mathbf{YY}^T \\
& = & \frac{1}{n}(\mathbf{PX})(\mathbf{PX})^T \\
& = & \frac{1}{n}\mathbf{PXX}^{T}\mathbf{P}^{T} \\
& = & \mathbf{P}(\frac{1}{n}\mathbf{XX}^{T})\mathbf{P}^{T} \\
\mathbf{C_Y} & = & \mathbf{P}\mathbf{C_X P}^T
\end{eqnarray*}
Note that we have identified the covariance matrix of $\mathbf{X}$ in the last line.

Our plan is to recognize that any symmetric matrix $\mathbf{A}$ is diagonalized by an orthogonal matrix of its eigenvectors (by Theorems 3 and 4 from Appendix A). For a symmetric matrix $\mathbf{A}$ Theorem 4 provides $\mathbf{A}=\mathbf{EDE}^T$, where $\mathbf{D}$ is a diagonal matrix and $\mathbf{E}$ is a matrix of eigenvectors of $\mathbf{A}$ arranged as columns.\footnote{The matrix $\mathbf{A}$ might have $r\leq m$ orthonormal eigenvectors where $r$ is the rank of the matrix. When the rank of $\mathbf{A}$ is less than $m$, $\mathbf{A}$ is {\it degenerate} or all data occupy a subspace of dimension $r\leq m$. Maintaining the constraint of orthogonality, we can remedy this situation by selecting $(m-r)$ additional orthonormal vectors to ``fill up'' the matrix $\mathbf{E}$. These additional vectors do not effect the final solution because the variances associated with these directions are zero.}

Now comes the trick. {\it We select the matrix $\mathbf{P}$ to be a matrix where each row $\mathbf{p_i}$ is an eigenvector of $\frac{1}{n}\mathbf{XX}^T$.} By this selection, $\mathbf{P \equiv E^{T}}$. With this relation and Theorem 1 of Appendix A ($\mathbf{P}^{-1}=\mathbf{P}^{T}$) we can finish evaluating $\mathbf{C_Y}$.
\begin{eqnarray*}
\mathbf{C_Y} & = & \mathbf{PC_X P}^{T} \\
& = & \mathbf{P}(\mathbf{E}^{T}\mathbf{DE})\mathbf{P}^{T} \\
& = & \mathbf{P}(\mathbf{P}^{T}\mathbf{DP})\mathbf{P}^{T} \\
& = & (\mathbf{PP}^{T})\mathbf{D}(\mathbf{PP}^{T}) \\
& = & (\mathbf{PP}^{-1})\mathbf{D}(\mathbf{PP}^{-1}) \\
\mathbf{C_Y} & = & \mathbf{D}
\end{eqnarray*}
It is evident that the choice of $\mathbf{P}$ diagonalizes $\mathbf{C_Y}$. This was the goal for PCA. We can summarize the results of PCA in the matrices $\mathbf{P}$ and $\mathbf{C_Y}$.
\begin{itemize}
\item The principal components of $\mathbf{X}$ are the eigenvectors of $\mathbf{C_X} = \frac{1}{n}\mathbf{XX}^T$.
\item The $i^{th}$ diagonal value of $\mathbf{C_Y}$ is the variance of $\mathbf{X}$ along $\mathbf{p_i}$.
\end{itemize}
In practice computing PCA of a data set $\mathbf{X}$ entails (1) subtracting off the mean of each measurement type and (2) computing the eigenvectors of $\mathbf{C_X}$. This solution is 
demonstrated in Matlab code included in Appendix B.

\section{A More General Solution Using SVD}
This section is the most mathematically involved and can be skipped without much loss of continuity. It is presented solely for completeness. We derive another algebraic solution for PCA and in the process, find that PCA is closely related to singular value decomposition (SVD). In fact, the two are so intimately related that the names are often used interchangeably. What we will see though is that SVD is a more general method of understanding {\it change of basis}.

We begin by quickly deriving the decomposition. In the following section we interpret the decomposition and in the last section we relate these results to {\it PCA}.

\subsection{Singular Value Decomposition}
Let $\mathbf{X}$ be an arbitrary $n \times m$ matrix\footnote{Notice that in this section only we are reversing convention from $m \times n$ to $n \times m$. The reason for this derivation will become clear in section 6.3.} and $\mathbf{X}^T\mathbf{X}$ be a rank $r$, square, symmetric $m \times m$ matrix. In a seemingly unmotivated fashion, let us define all of the quantities of interest.
\begin{itemize}
\item $\{\mathbf{\hat{v}}_1,\mathbf{\hat{v}}_2,\ldots,\mathbf{\hat{v}}_r\}$ is the set of {\it orthonormal} $m \times 1$ eigenvectors with associated eigenvalues $\{\lambda_1,\lambda_2,\ldots,\lambda_r\}$ for the symmetric matrix $\mathbf{X}^T\mathbf{X}$.
$$(\mathbf{X}^{T}\mathbf{X})\mathbf{\hat{v}}_i = \lambda_i\mathbf{\hat{v}}_i$$
\item $\sigma_i \equiv \sqrt{\lambda_i}$ are positive real and termed the {\it singular values}.
\item $\{\mathbf{\hat{u}}_1,\mathbf{\hat{u}}_2,\ldots,\mathbf{\hat{u}}_r\}$ is the set of $n \times 1$ vectors defined by \mbox{$\mathbf{\mathbf{\hat{u}_i} \equiv \frac{1}{\sigma_i}X\hat{v}_i}$}.
\end{itemize}
The final definition includes two new and unexpected properties.
\begin{itemize}
\item $\mathbf{\hat{u}_i} \cdot \mathbf{\hat{u}_j} = \left\{\begin{tabular}{cl}1 & \;\;if $i = j$ \\ 0 & \;\;otherwise\\\end{tabular}\right.$
\item  \mbox{$\Vert\mathbf{X\hat{v}_i}\Vert = \sigma_i$}
\end{itemize}
These properties are both proven in Theorem 5. We now have all of the pieces to construct the decomposition. The scalar version of singular value decomposition is just a restatement of the third definition.
\begin{equation}
\mathbf{X\hat{v}}_i = \sigma_i\mathbf{\hat{u}}_i
\label{eqn:value-svd}
\end{equation}
This result says a quite a bit. $\mathbf{X}$ multiplied by an eigenvector of $\mathbf{X}^{T}\mathbf{X}$ is equal to a scalar times another vector. The set of eigenvectors \mbox{$\{\mathbf{\hat{v}}_1,\mathbf{\hat{v}}_2,\ldots,\mathbf{\hat{v}}_r\}$} and the set of vectors \mbox{$\{\mathbf{\hat{u}}_1,\mathbf{\hat{u}}_2,\ldots,\mathbf{\hat{u}}_r\}$} are both orthonormal sets or bases in $r$-dimensional space.

\begin{figure*}[t]
\begin{quote}{\sf 
The scalar form of SVD is expressed in equation~\ref{eqn:value-svd}.
\[\mathbf{X\hat{v}}_i = \sigma_i\mathbf{\hat{u}}_i\]
The mathematical intuition behind the construction of the matrix form is that we want to express all $n$ scalar equations in just one equation. It is easiest to understand this process graphically. Drawing the matrices of equation~\ref{eqn:value-svd} looks likes the following.

\centerline{\includegraphics[width=0.45\textwidth]{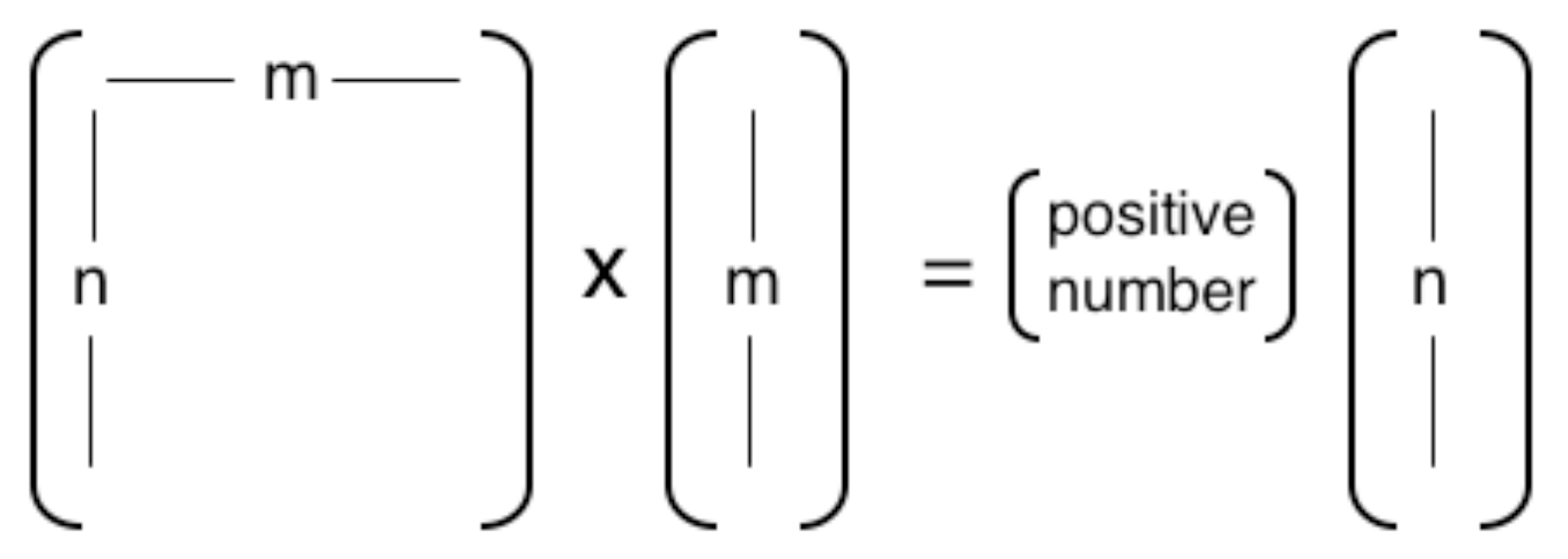}}

We can construct three new matrices $\mathbf{V}$, $\mathbf{U}$ and $\Sigma$. All singular values are first rank-ordered \mbox{$\sigma_{\tilde{1}} \geq \sigma_{\tilde{2}} \geq \ldots \geq \sigma_{\tilde{r}}$}, and the corresponding vectors are indexed in the same rank order. Each pair of associated vectors $\mathbf{\hat{v}_i}$ and $\mathbf{\hat{u}_i}$ is stacked in the $i^{th}$ column along their respective matrices. The corresponding singular value $\sigma_i$ is placed along the diagonal (the $ii^{th}$ position) of $\Sigma$. This generates the equation $\mathbf{XV=U}\Sigma$, which looks like the following.

\centerline{\includegraphics[width=0.75\textwidth]{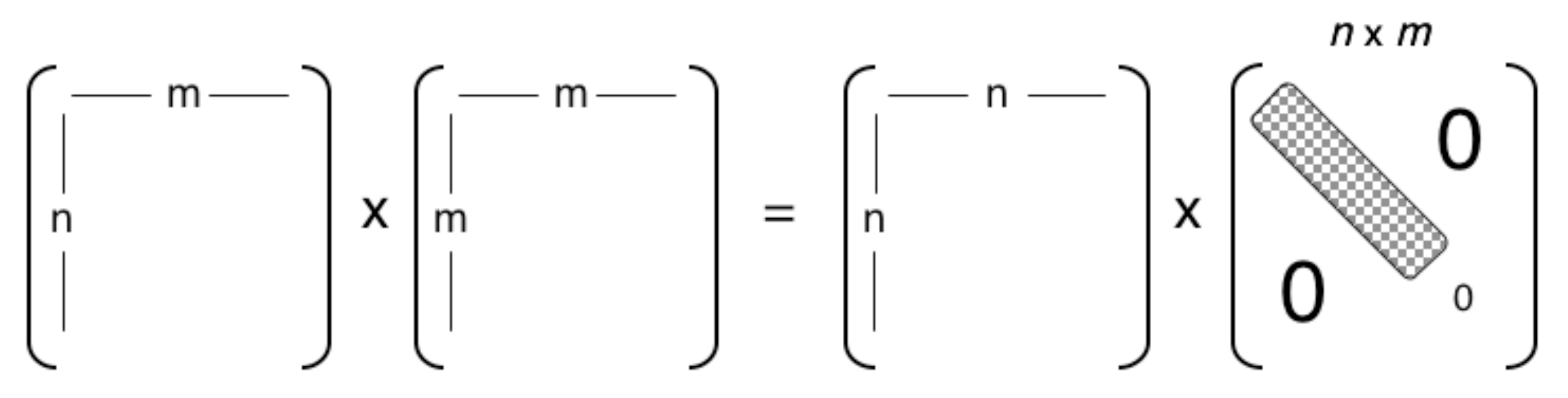}}

The matrices $\mathbf{V}$ and $\mathbf{U}$ are $m \times m$ and $n \times n$ matrices respectively and $\Sigma$ is a diagonal matrix with a few non-zero values (represented by the checkerboard) along its diagonal. Solving this single matrix equation solves all $n$ ``value'' form equations.
}\end{quote}
\caption{Construction of the matrix form of SVD (Equation~\ref{eqn:svd-matrix}) from the scalar form (Equation~\ref{eqn:value-svd}).}
\label{diagram:svd-construction}
\end{figure*}

We can summarize this result for all vectors in one matrix multiplication by following the prescribed construction in Figure~\ref{diagram:svd-construction}. We start by constructing a new diagonal matrix $\Sigma$.
$$\Sigma \equiv \left[ \begin{array}{cccccc} \sigma_{\tilde{1}} & & & & & \\ & \ddots & & & \mbox{{\Huge 0}} & \\ & & \sigma_{\tilde{r}} & & & \\ & & & 0 & & \\ & \mbox{{\Huge 0}} & & & \ddots & \\ & & & & & 0 \end{array} \right]$$
where \mbox{$\sigma_{\tilde{1}} \geq \sigma_{\tilde{2}} \geq \ldots \geq \sigma_{\tilde{r}}$} are the rank-ordered set of singular values. Likewise we construct accompanying orthogonal matrices,
\begin{eqnarray*}
\mathbf{V} & = & \left[\mathbf{\hat{v}_{\tilde{1}}}\;\mathbf{\hat{v}_{\tilde{2}}}\;\ldots\;\mathbf{\hat{v}_{\tilde{m}}} \right] \\
\mathbf{U} & = & \left[\mathbf{\hat{u}_{\tilde{1}}}\;\mathbf{\hat{u}_{\tilde{2}}}\;\ldots\;\mathbf{\hat{u}_{\tilde{n}}} \right] 
\end{eqnarray*}
where we have appended an additional $(m-r)$ and $(n-r)$ orthonormal vectors to ``fill up'' the matrices for $\mathbf{V}$ and $\mathbf{U}$ respectively (i.e. to deal with degeneracy issues). Figure~\ref{diagram:svd-construction} provides a graphical representation of how all of the pieces fit together to form the matrix version of {\it SVD}.
$$\mathbf{XV = U} \Sigma$$
where each column of $\mathbf{V}$ and $\mathbf{U}$ perform the scalar version of the decomposition (Equation~\ref{eqn:value-svd}). Because $\mathbf{V}$ is orthogonal, we can multiply both sides by $\mathbf{V}^{-1}=\mathbf{V}^{T}$ to arrive at the final form of the decomposition.
\begin{equation}
\mathbf{X = U} \Sigma \mathbf{V}^T
\label{eqn:svd-matrix}
\end{equation}
Although derived without motivation, this decomposition is quite powerful. Equation~\ref{eqn:svd-matrix} states that {\it any} arbitrary matrix $\mathbf{X}$ can be converted to an orthogonal matrix, a diagonal matrix and another orthogonal matrix (or a rotation, a stretch and a second rotation). Making sense of Equation~\ref{eqn:svd-matrix} is the subject of the next section.

\subsection{Interpreting SVD}
The final form of SVD is a concise but thick statement. Instead let us reinterpret Equation~\ref{eqn:value-svd} as
\[\mathbf{Xa} = k\mathbf{b}\]
where $\mathbf{a}$ and $\mathbf{b}$ are column vectors and $k$ is a scalar constant. The set \mbox{$\{\mathbf{\hat{v}_1},\mathbf{\hat{v}_2},\ldots,\mathbf{\hat{v}_m}\}$} is analogous to $\mathbf{a}$ and the set \mbox{$\{\mathbf{\hat{u}_1},\mathbf{\hat{u}_2},\ldots,\mathbf{\hat{u}_n}\}$} is analogous to $\mathbf{b}$. What is unique though is that \mbox{$\{\mathbf{\hat{v}_1},\mathbf{\hat{v}_2},\ldots,\mathbf{\hat{v}_m}\}$} and \mbox{$\{\mathbf{\hat{u}_1},\mathbf{\hat{u}_2},\ldots,\mathbf{\hat{u}_n}\}$} are orthonormal sets of vectors which {\it span} an $m$ or $n$ dimensional space, respectively. In particular, loosely speaking these sets appear to span all possible ``inputs'' (i.e. $\mathbf{a}$) and ``outputs'' (i.e. $\mathbf{b}$). Can we formalize the view that \mbox{$\{\mathbf{\hat{v}_1},\mathbf{\hat{v}_2},\ldots,\mathbf{\hat{v}_n}\}$} and \mbox{$\{\mathbf{\hat{u}_1},\mathbf{\hat{u}_2},\ldots,\mathbf{\hat{u}_n}\}$} span all possible ``inputs'' and ``outputs''?

We can manipulate Equation~\ref{eqn:svd-matrix} to make this fuzzy hypothesis more precise.
\begin{eqnarray*}
\mathbf{X} & = & \mathbf{U}\Sigma\mathbf{V}^{T} \\
\mathbf{U}^{T}\mathbf{X} & = & \Sigma\mathbf{V}^{T} \\
\mathbf{U}^{T}\mathbf{X} & = & \mathbf{Z}
\end{eqnarray*}
where we have defined $\mathbf{Z }\equiv \Sigma \mathbf{V}^{T}$. Note that the previous columns \mbox{$\{\mathbf{\hat{u}_1},\mathbf{\hat{u}_2},\ldots,\mathbf{\hat{u}_n}\}$} are now rows in $\mathbf{U}^T$. Comparing this equation to Equation~\ref{eqn:basis-transform}, \mbox{$\{\mathbf{\hat{u}_1},\mathbf{\hat{u}_2},\ldots,\mathbf{\hat{u}_n}\}$} perform the same role as \mbox{$\{\mathbf{\hat{p}_1},\mathbf{\hat{p}_2},\ldots,\mathbf{\hat{p}_m}\}$}. Hence, $\mathbf{U}^T$ is a {\it change of basis} from $\mathbf{X}$ to $\mathbf{Z}$. Just as before, we were transforming column vectors, we can again infer that we are transforming column vectors. The fact that the orthonormal basis $\mathbf{U}^T$ (or $\mathbf{P}$) transforms column vectors means that $\mathbf{U}^T$ is a basis that spans the columns of $\mathbf{X}$. Bases that span the columns are termed the {\it column space} of $\mathbf{X}$. The column space formalizes the notion of what are the possible ``outputs'' of any matrix.

There is a funny symmetry to SVD such that we can define a similar quantity - the {\it row space}.
\begin{eqnarray*}
\mathbf{XV} & = & \Sigma\mathbf{U} \\
(\mathbf{XV})^{T} & = & (\Sigma\mathbf{U})^{T} \\
\mathbf{V}^{T}\mathbf{X}^{T} & = & \mathbf{U}^{T} \Sigma \\
\mathbf{V}^{T}\mathbf{X}^{T} & = & \mathbf{Z}
\end{eqnarray*}
where we have defined $\mathbf{Z \equiv U^{T}}\Sigma$. Again the rows of $\mathbf{V}^T$ (or the columns of $\mathbf{V}$) are an orthonormal basis for transforming $\mathbf{X}^T$ into $\mathbf{Z}$. Because of the transpose on $\mathbf{X}$, it follows that $\mathbf{V}$ is an orthonormal basis spanning the {\it row space} of $\mathbf{X}$. The row space likewise formalizes the notion of what are possible ``inputs'' into an arbitrary matrix.

We are only scratching the surface for understanding the full implications of SVD. For the purposes of this tutorial though, we have enough information to understand how PCA will fall within this framework.

\subsection{SVD and PCA}
It is evident that PCA and SVD are intimately related. Let us return to the original $m \times n$ data matrix $\mathbf{X}$. We can define a new matrix $\mathbf{Y}$ as an $n \times m$ matrix.\footnote{$\mathbf{Y}$ is of the appropriate $n \times m$ dimensions laid out in the derivation of section 6.1. This is the reason for the ``flipping'' of dimensions in 6.1 and Figure 4.}
$$\mathbf{Y} \equiv \frac{1}{\sqrt{n}}\mathbf{X}^T$$
where each {\it column} of $\mathbf{Y}$ has zero mean. The choice of $\mathbf{Y}$ becomes clear by analyzing $\mathbf{Y}^T\mathbf{Y}$.
\begin{eqnarray*}
\mathbf{Y}^T\mathbf{Y} & = & \left(\frac{1}{\sqrt{n}}\mathbf{X}^T\right)^{T}\left(\frac{1}{\sqrt{n}}\mathbf{X}^T\right) \\
& = & \frac{1}{n}\mathbf{XX}^{T} \\
\mathbf{Y}^{T}\mathbf{Y} & = & \mathbf{C_X}
\end{eqnarray*}
By construction $\mathbf{Y}^T\mathbf{Y}$ equals the covariance matrix of $\mathbf{X}$. From section 5 we know that the principal components of $\mathbf{X}$ are the eigenvectors of $\mathbf{C_X}$. If we calculate the SVD of $\mathbf{Y}$, the columns of matrix $\mathbf{V}$ contain the eigenvectors of $\mathbf{Y}^T\mathbf{Y} = \mathbf{C_X}$. {\it Therefore, the columns of $\mathbf{V}$ are the principal components of $\mathbf{X}$}. This second algorithm is encapsulated in Matlab code included in Appendix B.

What does this mean? $\mathbf{V}$ spans the row space of $\mathbf{Y} \equiv \frac{1}{\sqrt{n}}\mathbf{X}^T$. Therefore, $\mathbf{V}$ must also span the column space of $\frac{1}{\sqrt{n}}\mathbf{X}$. We can conclude that finding the principal components amounts to finding an orthonormal basis that spans the {\it column space} of $\mathbf{X}$.\footnote{If the final goal is to find an orthonormal basis for the coulmn space of $\mathbf{X}$ then we can calculate it directly without constructing $\mathbf{Y}$. By symmetry the columns of $\mathbf{U}$ produced by the SVD of $\frac{1}{\sqrt{n}}\mathbf{X}$ must also be the principal components.} 

\section{Discussion}

\begin{figure}
\framebox{\parbox{3.2in}{
{\bf Quick Summary of PCA}
{\sf
\begin{enumerate}
\item Organize data as an $m \times n$ matrix, where $m$ is the number of measurement types and $n$ is the number of samples.
\item Subtract off the mean for each measurement type.
\item Calculate the SVD or the eigenvectors of the covariance.
\end{enumerate}
}}}
\caption{A step-by-step instruction list on how to perform principal component analysis} 
\label{fig:summary}
\end{figure}

Principal component analysis (PCA) has widespread applications because it reveals simple underlying structures in complex data sets using analytical solutions from linear algebra. Figure~\ref{fig:summary} provides a brief summary for implementing PCA.

A primary benefit of PCA arises from quantifying the importance of each dimension for describing the variability of a data set. In particular, the measurement of the variance along each principle component provides a means for comparing the relative importance of each dimension. An implicit hope behind employing this method is that the variance along a small number of principal components (i.e. less than the number of measurement types) provides a reasonable characterization of the complete data set. This statement is the precise intuition behind any method of {\em dimensional reduction} -- a vast arena of active research. In the example of the spring, PCA identifies that a majority of variation exists along a single dimension (the direction of motion $\hat{\mathbf{x}}$), eventhough 6 dimensions are recorded.

Although PCA ``works'' on a multitude of real world problems, any diligent scientist or engineer must ask {\em when does PCA fail?} Before we answer this question, let us note a remarkable feature of this algorithm. PCA is completely {\em non-parametric}: any data set can be plugged in and an answer comes out, requiring no parameters to tweak and no regard for how the data was recorded. From one perspective, the fact that PCA is non-parametric (or plug-and-play) can be considered a positive feature because the answer is unique and independent of the user. From another perspective the fact that PCA is agnostic to the source of the data is also a weakness. For instance, consider tracking a person on a ferris wheel in Figure~\ref{fig:failures}a. The data points can be cleanly described by a single variable, the precession angle of the wheel $\theta$, however PCA would fail to recover this variable.

\subsection{Limits and Statistics of Dimensional Reduction}

\begin{figure}[t]
\includegraphics[width=0.47\textwidth]{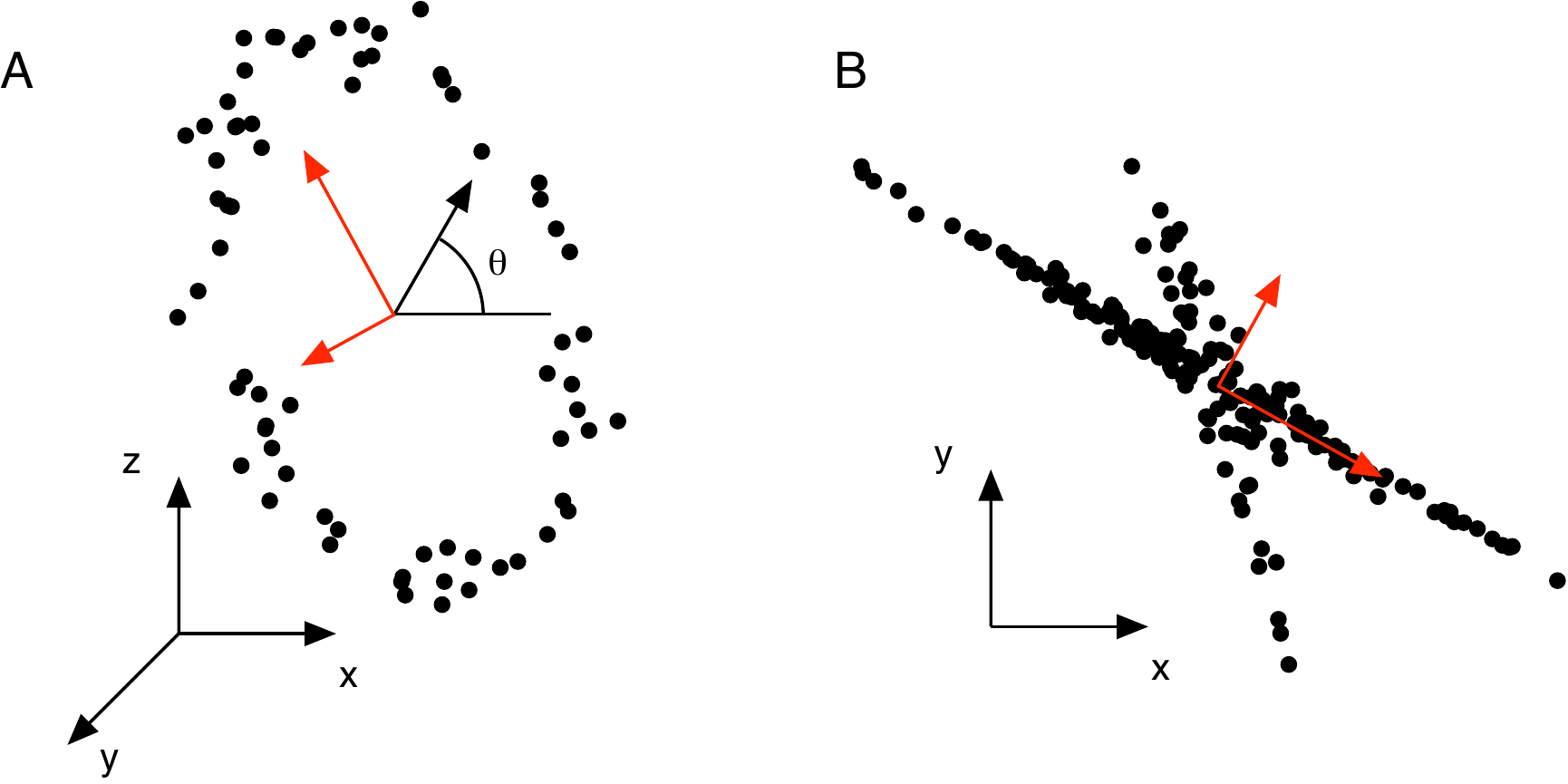}
\caption{Example of when PCA fails (red lines). (a) Tracking a person on a ferris wheel (black dots). All dynamics can be described by the phase of the wheel $\theta$, a non-linear combination of the naive basis. (b) In this example data set, non-Gaussian distributed data and non-orthogonal axes causes PCA to fail. The axes with the largest variance do not correspond to the appropriate answer.}
\label{fig:failures}
\end{figure}

A deeper appreciation of the limits of PCA requires some consideration about the underlying assumptions and in tandem, a more rigorous description of the source of data. Generally speaking, the primary motivation behind this method is to decorrelate the data set, i.e. remove second-order dependencies. The manner of approaching this goal is loosely akin to how one might explore a town in the Western United States: drive down the longest road running through the town. When one sees another big road, turn left or right and drive down this road, and so forth. In this analogy, PCA requires that each new road explored must be perpendicular to the previous, but clearly this requirement is overly stringent and the data (or town) might be arranged along non-orthogonal axes, such as Figure~\ref{fig:failures}b. Figure~\ref{fig:failures} provides two examples of this type of data where PCA provides unsatisfying results.

To address these problems, we must define what we consider optimal results. In the context of dimensional reduction, one measure of success is the degree to which a reduced representation can predict the original data. In statistical terms, we must define an error function (or loss function). It can be proved that under a common loss function, mean squared error (i.e. $L_2$ norm), PCA provides the optimal reduced representation of the data. This means that selecting orthogonal directions for principal components is the best solution to predicting the original data. Given the examples of Figure~\ref{fig:failures}, how could this statement be true? Our intuitions from Figure~\ref{fig:failures} suggest that this result is somehow misleading.

The solution to this paradox lies in the goal we selected for the analysis. The goal of the analysis is to decorrelate the data, or said in other terms, the goal is to remove second-order dependencies in the data. In the data sets of Figure~\ref{fig:failures}, higher order dependencies exist between the variables. Therefore, removing second-order dependencies is insufficient at revealing all structure in the data.\footnote{When are second order dependencies sufficient for revealing all dependencies in a data set? This statistical condition is met when the first and second order statistics are {\em sufficient statistics} of the data. This occurs, for instance, when a data set is Gaussian distributed.}

Multiple solutions exist for removing higher-order dependencies. For instance, if prior knowledge is known about the problem, then a nonlinearity (i.e. {\em kernel}) might be applied to the data to transform the data to a more appropriate naive basis. For instance, in Figure~\ref{fig:failures}a, one might examine the polar coordinate representation of the data. This parametric approach is often termed {\it kernel PCA}.

Another direction is to impose more general statistical definitions of dependency within a data set, e.g. requiring that data along reduced dimensions be {\em statistically independent}. This class of algorithms, termed, {\em independent component analysis} (ICA), has been demonstrated to succeed in many domains where PCA fails. ICA has been applied to many areas of signal and image processing, but suffers from the fact that solutions are (sometimes) difficult to compute.

Writing this paper has been an extremely instructional experience for me. I hope that this paper helps to demystify the motivation and results of PCA, and the underlying assumptions behind this important analysis technique. Please send me a note if this has been useful to you as it inspires me to keep writing!

\appendix

\section{Linear Algebra}
This section proves a few unapparent theorems in linear algebra, which are crucial to this paper. \\

{\bf1. The inverse of an orthogonal matrix is its transpose.} \\

Let $\mathbf{A}$ be an $m \times n$ orthogonal matrix where $\mathbf{a_i}$ is the $i^{th}$ column vector. The $ij^{th}$ element of $\mathbf{A}^{T}\mathbf{A}$ is 
$$(\mathbf{A}^{T}\mathbf{A})_{ij} = \mathbf{a_{i}}^{T}\mathbf{a_{j}} =  \left\{ 
	\begin{array}{ll}
		1 & if \;\;i=j \\
		0 & otherwise \\
	\end{array} \right.$$
Therefore, because $\mathbf{A}^{T}\mathbf{A}=\mathbf{I}$, it follows that $\mathbf{A}^{-1}=\mathbf{A}^{T}$. \\

{\bf 2. For any matrix $\mathbf{A}$, $\mathbf{A}^{T}\mathbf{A}$ and $\mathbf{AA}^{T}$ are symmetric.} \\
\begin{eqnarray*}
(\mathbf{AA}^{T})^T & = & \mathbf{A}^{TT}\mathbf{A}^{T} = \mathbf{AA}^{T} \\
(\mathbf{A}^{T}\mathbf{A})^T & = & \mathbf{A}^{T}\mathbf{A}^{TT} = \mathbf{A}^{T}\mathbf{A}
\end{eqnarray*}
{\bf 3. A matrix is symmetric if and only if it is orthogonally diagonalizable.} \\

Because this statement is bi-directional, it requires a two-part ``if-and-only-if'' proof. One needs to prove the forward and the backwards ``if-then'' cases.

Let us start with the forward case. If $\mathbf{A}$ is orthogonally diagonalizable, then $\mathbf{A}$ is a symmetric matrix. By hypothesis, orthogonally diagonalizable means that there exists some $\mathbf{E}$  such that $\mathbf{A}=\mathbf{EDE}^{T}$, where $\mathbf{D}$ is a diagonal matrix and $\mathbf{E}$ is some special matrix which diagonalizes $\mathbf{A}$. Let us compute $\mathbf{A}^{T}$.
$$\mathbf{A}^T = (\mathbf{EDE}^{T})^{T} = \mathbf{E}^{TT}\mathbf{D}^{T}\mathbf{E}^{T} = \mathbf{EDE}^{T} = \mathbf{A}$$

Evidently, if $\mathbf{A}$ is orthogonally diagonalizable, it must also be symmetric. 

The reverse case is more involved and less clean so it will be left to the reader. In lieu of this, hopefully the ``forward'' case is suggestive if not somewhat convincing. \\

{\bf 4. A symmetric matrix is diagonalized by a matrix of its orthonormal eigenvectors.} \\

Let $\mathbf{A}$ be a square \mbox{$n \times n$} symmetric matrix with associated eigenvectors $\{\mathbf{e_1, e_2, \ldots, e_n} \}$. Let $ \mathbf{E}=[\mathbf{e_1\;e_2\;\ldots\;e_n]}$ where the $i^{th}$ column of $\mathbf{E}$ is the eigenvector $\mathbf{e_i}$. This theorem asserts that there exists a diagonal matrix $\mathbf{D}$ such that $\mathbf{A}=\mathbf{EDE}^{T}$.

This proof is in two parts. In the first part, we see that the any matrix can be orthogonally diagonalized if and only if it that matrix's eigenvectors are all linearly independent. In the second part of the proof, we see that a symmetric matrix has the special property that all of its eigenvectors are not just linearly independent but also orthogonal, thus completing our proof.

In the first part of the proof, let $\mathbf{A}$ be just some matrix, not necessarily symmetric, and let it have independent eigenvectors (i.e. no degeneracy). Furthermore, let $ \mathbf{E}=[\mathbf{e_1\;e_2\;\ldots\;e_n]}$ be the matrix of eigenvectors placed in the columns. Let $\mathbf{D}$ be a diagonal matrix where the $i^{th}$ eigenvalue is placed in the $ii^{th}$ position.

We will now show that $\mathbf{AE=ED}$. We can examine the columns of the right-hand and left-hand sides of the equation.
\begin{displaymath}
\begin{array}{rrcl}
	\mathsf{Left\;hand\;side:} & \mathbf{AE} & = & [\mathbf{Ae_1}\;\mathbf{Ae_2}\;\ldots\;\mathbf{Ae_n}] \\
	\mathsf{Right\;hand\;side:} & \mathbf{ED} & = & [\lambda_{1}\mathbf{e_1}\:\lambda_{2}\mathbf{e_2}\:\ldots\:\lambda_{n}\mathbf{e_n}]
\end{array}
\end{displaymath}
Evidently, if $\mathbf{AE=ED}$ then $\mathbf{Ae_i}=\lambda_{i}\mathbf{e_i}$ for all $i$. This equation is the definition of the eigenvalue equation. Therefore, it must be that $\mathbf{AE=ED}$. A little rearrangement provides $\mathbf{A=EDE}^{-1}$, completing the first part the proof.

For the second part of the proof, we show that a symmetric matrix always has orthogonal eigenvectors. For some symmetric matrix, let $\lambda_{1}$ and $\lambda_{2}$ be distinct eigenvalues for eigenvectors $\mathbf{e_1}$ and $\mathbf{e_2}$.
\begin{eqnarray*}
	\lambda_1\mathbf{e_1}\cdot\mathbf{e_2} & = & (\lambda_1 \mathbf{e_1})^{T} \mathbf{e_2} \\
	& = & (\mathbf{Ae_1})^{T} \mathbf{e_2} \\
	& = & \mathbf{e_1}^{T}\mathbf{A}^{T} \mathbf{e_2} \\
	& = & \mathbf{e_1}^{T}\mathbf{A} \mathbf{e_2} \\
	& = & \mathbf{e_1}^{T} (\lambda_{2}\mathbf{e_2}) \\
	\lambda_1\mathbf{e_1}\cdot\mathbf{e_2} & = & \lambda_2\mathbf{e_1}\cdot\mathbf{e_2}
\end{eqnarray*}
By the last relation we can equate that \mbox{$ (\lambda_1-\lambda_2)\mathbf{e_1}\cdot\mathbf{e_2} = 0$}. Since we have conjectured that the eigenvalues are in fact unique, it must be the case that $\mathbf{e_1}\cdot\mathbf{e_2} = 0$. Therefore, the eigenvectors of a symmetric matrix are orthogonal.

Let us back up now to our original postulate that $\mathbf{A}$ is a symmetric matrix. By the second part of the proof, we know that the eigenvectors of $\mathbf{A}$ are all orthonormal (we choose the eigenvectors to be normalized). This means that $\mathbf{E}$ is an orthogonal matrix so by theorem 1, $\mathbf{E}^T=\mathbf{E}^{-1}$ and we can rewrite the final result.
\[\mathbf{A=EDE}^T\].
Thus, a symmetric matrix is diagonalized by a matrix of its eigenvectors.\\

{\bf 5. For any arbitrary $m \times n$ matrix $\mathbf{X}$, the symmetric matrix $\mathbf{X}^{T}\mathbf{X}$ has a set of orthonormal eigenvectors of $\{\mathbf{\hat{v}_1,\hat{v}_2,\ldots,\hat{v}_n}\}$ and a set of associated eigenvalues $\{\mathbf{\lambda_1,\lambda_2,\ldots,\lambda_n}\}$. The set of vectors $\{\mathbf{X}\mathbf{\hat{v}_1},\mathbf{X}\mathbf{\hat{v}_2},\ldots,\mathbf{X}\mathbf{\hat{v}_n}\}$ then form an orthogonal basis, where each vector $\mathbf{X}\mathbf{\hat{v}_i}$ is of length $\sqrt{\lambda_i}$.} \\

All of these properties arise from the dot product of any two vectors from this set.
\begin{eqnarray*}
(\mathbf{X\hat{v}_i})\cdot(\mathbf{X\hat{v}_j}) & = & (\mathbf{X\hat{v}_i})^{T}(\mathbf{X\hat{v}_j}) \\
& = & \mathbf{\hat{v}_i}^{T}\mathbf{X}^{T}\mathbf{X}\mathbf{\hat{v}_j} \\ 
& = & \mathbf{\hat{v}_i}^{T}(\lambda_j\mathbf{\hat{v}_j}) \\
& = & \lambda_j\mathbf{\hat{v}_i}\cdot\mathbf{\hat{v}_j} \\
(\mathbf{X\hat{v}_i})\cdot(\mathbf{X\hat{v}_j}) & = & \lambda_j\delta_{ij} \\
\end{eqnarray*} 
The last relation arises because the set of eigenvectors of $\mathbf{X}$ is orthogonal resulting in the Kronecker delta. In more simpler terms the last relation states:
\begin{displaymath}
(\mathbf{X\hat{v}_i})\cdot(\mathbf{X\hat{v}_j}) = \left\{ 
	\begin{array}{ll}
		\lambda_j & i=j \\
		0 & i \neq j \\
	\end{array} \right.
\end{displaymath}
This equation states that any two vectors in the set are orthogonal.

The second property arises from the above equation by realizing that the length squared of each vector is defined as:
$$\|\mathbf{X\hat{v}_i}\|^2 = (\mathbf{X\hat{v}_i})\cdot(\mathbf{X\hat{v}_i}) = \lambda_i$$

\section{Code}

This code is written for Matlab 6.5 (Release 13) from Mathworks\footnote{{\tt http://www.mathworks.com}}. The code is not computationally efficient but explanatory (terse comments begin with a \%).
\linebreak\linebreak
This first version follows Section 5 by examining the covariance of the data set.

\input{pca1.m}

This second version follows section 6 computing PCA through SVD.

\input{pca2.m}

\end{document}